\documentclass[11pt,a4paper,table]{article}
\usepackage{authblk}
\usepackage[hyperref]{emnlp-ijcnlp-2019}
\usepackage{times}
\usepackage{latexsym}

\usepackage[utf8]{inputenc}

\usepackage{url}
\usepackage{calc}
\usepackage{xcolor}
\usepackage{graphicx}
\usepackage{multirow}
\usepackage{amsthm}
\usepackage{amssymb}
\usepackage{amsmath}
\usepackage{amsfonts}
\usepackage{color}
\usepackage{mdwlist}
\usepackage{microtype}
\usepackage{tabularx}
\usepackage{booktabs}
\usepackage{subcaption}
\usepackage{covington}
\usepackage{makecell}
\usepackage{arydshln}

\definecolor{Gray}{gray}{0.9}
\definecolor{LightGray}{gray}{0.88}
\definecolor{LighterGray}{gray}{0.92}

\usepackage{enumitem,amssymb}
\usepackage[russian,english]{babel}
\usepackage{CJKutf8}

\usepackage{float}

\aclfinalcopy 

\title{PolyResponse: A Rank-based Approach to Task-Oriented Dialogue with Application in Restaurant Search and Booking}

\author{\bf Matthew Henderson, Ivan Vuli\'{c}, I\~{n}igo Casanueva, Pawe\l{} Budzianowski, Daniela Gerz, \\  {\bf Sam Coope, Georgios Spithourakis, Tsung-Hsien Wen, Nikola Mrk\v{s}i\'{c}, Pei-Hao Su} \\ \\ PolyAI Limited, London, UK \\
\href{https://poly-ai.com/}{\small \texttt{poly-ai.com}}}

\begin{document}
\maketitle

\begin{abstract}
We present PolyResponse, a conversational search engine that supports task-oriented dialogue. It is a retrieval-based approach that bypasses the complex multi-component design of traditional task-oriented dialogue systems and the use of explicit semantics in the form of task-specific ontologies. The PolyResponse engine is trained on hundreds of millions of examples extracted from real conversations: it learns what responses are appropriate in different conversational contexts. It then ranks a large index of text and visual responses according to their similarity to the given context, and narrows down the list of relevant entities during the multi-turn conversation. We introduce a restaurant search and booking system powered by the PolyResponse engine, currently available in 8 different languages.
\end{abstract}

\section{Introduction and Background}
\label{s:introduction}
Task-oriented dialogue systems are primarily designed to search and interact with large databases which contain information pertaining to a certain \textit{dialogue domain}: the main purpose of such systems is to assist the users in accomplishing a well-defined task such as flight booking \cite{ElAsri:2017sigdial}, tourist information \cite{Henderson:14b}, restaurant search \cite{Williams:12}, or booking a taxi \cite{Budzianowski:2018emnlp}. These systems are typically constructed around rigid task-specific ontologies \cite{Henderson:14b,Mrksic:15} which enumerate the constraints the users can express using a collection of slots (e.g., \textsc{price range} for restaurant search) and their slot values (e.g., \textsc{cheap}, \textsc{expensive} for the aforementioned slots). Conversations are then modelled as a sequence of actions that constrain slots to particular values. This \textit{explicit} semantic space is manually engineered by the system designer. It serves as the output of the natural language understanding component as well as the input to the language generation component both in traditional modular systems \cite{young:10b,Eric:2017sigdial} and in more recent end-to-end task-oriented dialogue systems \cite[\textit{inter alia}]{Wen:17,Li:2017ijcnlp,Bordes:2017iclr,Budzianowski:2018emnlp}.


Working with such explicit semantics for task-oriented dialogue systems poses several critical challenges on top of the manual time-consuming domain ontology design. First, it is difficult to collect domain-specific data labelled with explicit semantic representations. As a consequence, despite recent data collection efforts to enable training of task-oriented systems across multiple domains \cite{ElAsri:2017sigdial,Budzianowski:2018emnlp}, annotated datasets are still few and far between, as well as limited in size and the number of domains covered.\footnote{For instance, the recently published MultiWOZ dataset \cite{Budzianowski:2018emnlp} contains a total of 115,424 dialogue turns scattered over 7 target domains. Other standard task-based datasets are typically single-domain and by several orders of magnitude smaller: DSTC2 \cite{Henderson:14b} contains 23,354 turns, Frames \cite{ElAsri:2017sigdial} 19,986 turns, and M2M \cite{Shah:2018naacl} spans 14,796 turns. On the other hand, the Reddit corpus which supports our system comprises 3.7B comments spanning a multitude of topics, divided into 256M (Reddit) conversational threads and generating 727M \textit{context-reply} pairs.} Second, the current approach constrains the types of dialogue the system can support, resulting in artificial conversations, and breakdowns when the user does not understand what the system can and cannot support. In other words, training a task-based dialogue system for voice-controlled search in a new domain always implies the complex, expensive, and time-consuming process of collecting and annotating sufficient amounts of in-domain dialogue data.

In this paper, we present a demo system based on an alternative approach to task-oriented dialogue. Relying on \textit{non-generative response retrieval} we describe the \textit{PolyResponse} conversational search engine and its application in the task of \textit{restaurant search and booking}. The engine is trained on hundreds of millions of real conversations from a general domain (i.e., Reddit), using an implicit representation of semantics that directly optimizes the task at hand. It learns what responses are appropriate in different conversational contexts, and consequently ranks a large pool of responses according to their relevance to the current user utterance and previous dialogue history (i.e., dialogue context). 

The technical aspects of the underlying conversational search engine are explained in detail in our recent work \cite{Henderson:2019acl}, while the details concerning the Reddit training data are also available in another recent publication \cite{Henderson:2019convai}. In this demo, we put focus on the actual practical usefulness of the search engine by demonstrating its potential in the task of {restaurant search}, and extending it to deal with multi-modal data. We describe a PolyReponse system that assists the users in finding a relevant restaurant according to their preference, and then additionally helps them to make a booking in the selected restaurant. Due to its retrieval-based design, with the PolyResponse engine there is no need to engineer a structured ontology, or to solve the difficult task of general language generation. This design also bypasses the construction of dedicated decision-making policy modules. The large ranking model already encapsulates a lot of knowledge about natural language and conversational flow.

Since retrieved system responses are presented visually to the user, the PolyResponse restaurant search engine is able to combine text responses with relevant visual information (e.g., photos from social media associated with the current restaurant and related to the user utterance), effectively yielding a \textit{multi-modal response}. This setup of using voice as input, and responding visually is becoming more and more prevalent with the rise of smart screens like Echo Show and even mixed reality. Finally, the PolyResponse restaurant search engine is \textit{multilingual}: it is currently deployed in 8 languages enabling search over restaurants in 8 cities around the world. System snapshots in four different languages are presented in Figure~\ref{fig:snapshots}, while screencast videos that illustrate the dialogue flow with the PolyResponse engine are available at: {\small \url{https://tinyurl.com/y3evkcfz}}.

\section{PolyResponse: Conversational Search}
\label{s:model}
\begin{figure}[!t]
\centering
\includegraphics[width=0.85\linewidth]{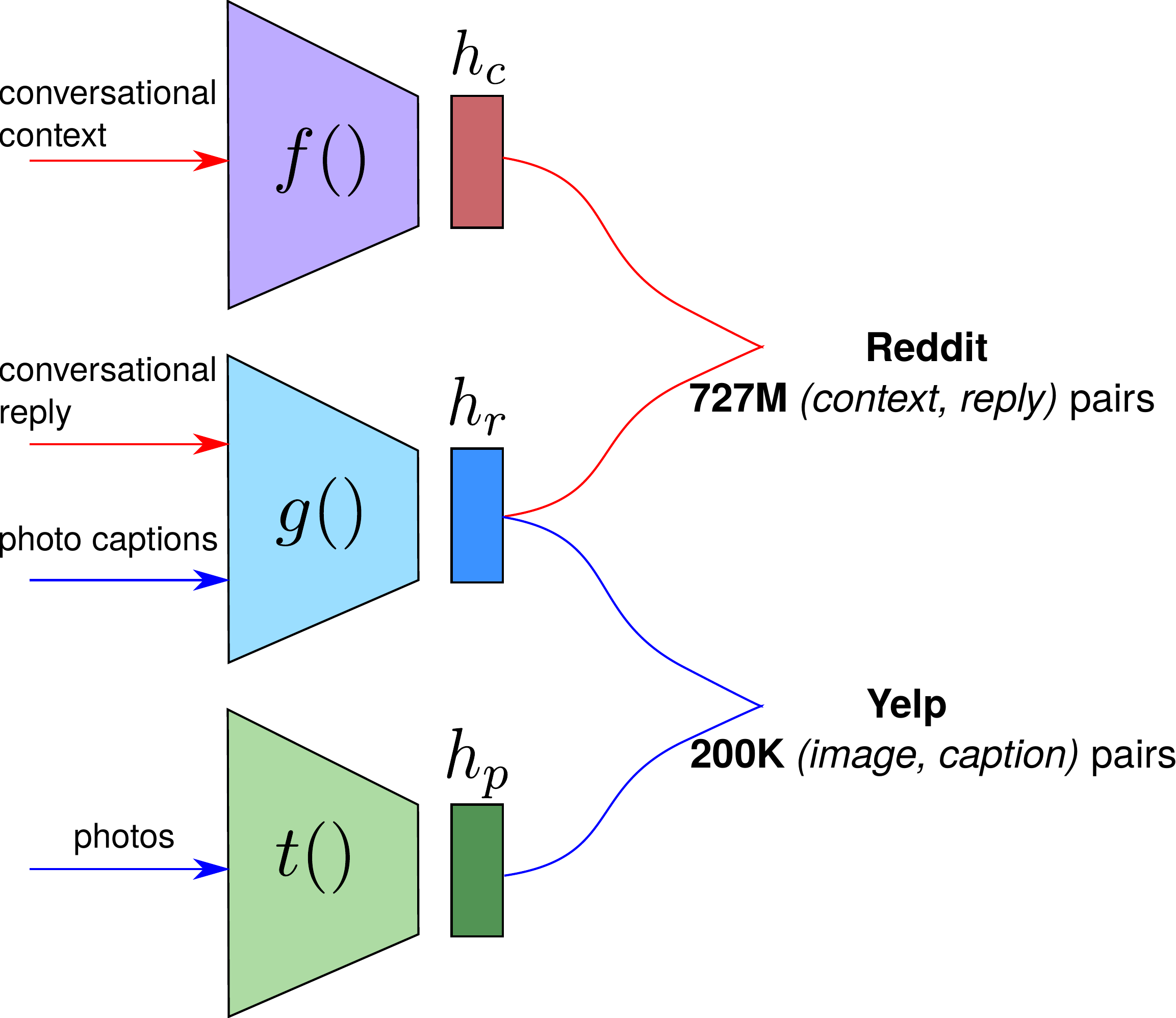}
\vspace{-1.5mm}
\caption{The PolyResponse ranking model: it encodes conversational contexts, replies, and photos to respective vectors $h_c$, $h_r$, and $h_p$.}
\vspace{-12.5mm}
\label{fig:arch}
\end{figure}


The PolyResponse system is powered by a single large conversational search engine, trained on a large amount of conversational and image data, as shown in Figure~\ref{fig:arch}. In simple words, it is a ranking model that learns to score conversational replies and images in a given conversational context. The highest-scoring responses are then retrieved as system outputs. The system computes two sets of similarity scores: \textbf{1)} $S(r,c)$ is the score of a candidate reply $r$ given a conversational context $c$, and \textbf{2)} $S(p,c)$ is the score of a candidate photo $p$ given a conversational context $c$. These scores are computed as a scaled cosine similarity of a vector that represents the
context ($h_c$), and a vector that represents the candidate response: a text reply ($h_r$) or a photo ($h_p$). For instance, $S(r,c)$ is computed as $S(r,c)=C cos(h_r,h_c)$, where $C$ is a learned constant. The part of the model dealing with text input (i.e., obtaining the encodings $h_c$ and $h_r$) follows the architecture introduced recently by \newcite{Henderson:2019acl}. We provide only a brief recap here; see the original paper for further details.


\paragraph{Text Representation.} The model, implemented as a deep neural network, learns to respond by training on hundreds of millions \textit{context-reply} $(c,r)$ pairs. First, similar to \newcite{Henderson:2017arxiv}, raw text from both $c$ and $r$ is converted to unigrams and bigrams. All input text is first lower-cased and tokenised, numbers with 5 or more digits get their digits replaced by a wildcard symbol \textit{\#}, while words longer than 16 characters are replaced by a wildcard token LONGWORD. Sentence boundary tokens are added to each sentence. The vocabulary consists of the unigrams that occur at least 10 times in a random 10M subset of the Reddit training set (see Figure~\ref{fig:arch}) plus the 200K most frequent bigrams in the same random subset.

During training, we obtain $d$-dimensional feature representations ($d=320$) shared between contexts and replies for each unigram and bigram jointly with other neural net parameters.\footnote{The model deals with out-of-vocabulary unigrams and bigrams by assigning a random id from 0 to 50,000 to each; this is then used to look up their embedding.} A state-of-the-art architecture based on transformers \cite{Vaswani:2017nips} is then applied to unigram and bigram vectors separately, which are then averaged to form the final $320$-dimensional encoding. That encoding is then passed through three fully-connected non-linear hidden layers of dimensionality $1,024$. The final layer is linear and maps the text into the final $l$-dimensional ($l=512$) representation: $h_c$ and $h_r$. Other standard and more sophisticated encoder models can also be used to provide final encodings $h_c$ and $h_r$, but the current architecture shows a good trade-off between speed and efficacy with strong and robust performance in our empirical evaluations on the response retrieval task using Reddit \cite{AlRfou:2016arxiv}, OpenSubtitles \cite{Lison:2016lrec}, and AmazonQA \cite{Wan:2016icdm} conversational test data, see \cite{Henderson:2019convai} for further details.\footnote{The comparisons of performance in the response retrieval task are also available online at: \url{https://github.com/PolyAI-LDN/conversational-datasets/}.}

In training the constant $C$ is constrained to lie between 0 and $\sqrt{l}$.\footnote{It is initialised to a random value between 0.5 and 1, and invariably converges to $\sqrt{l}$ by the end of
training. Empirically, this helps with learning.} Following \newcite{Henderson:2017arxiv}, the scoring function in the training objective aims to maximise the similarity score of context-reply  pairs that go together, while minimising the score of random pairings: negative examples. Training proceeds via SGD with batches comprising 500 pairs (1 positive and 499 negatives).

\paragraph{Photo Representation.}
Photos are represented using convolutional neural net (CNN) models pretrained on ImageNet \cite{Deng:2009cvpr}. We use a MobileNet model with a depth multiplier of 1.4, and an input dimension of $224 \times 224$ pixels as in \cite{Howard:2017arxiv}.\footnote{The pretrained model downloaded from TensorFlow Slim.} This provides a $1,280 \times 1.4 = 1,792$-dimensional representation of a photo, which is then passed through a single hidden layer of dimensionality $1,024$ with ReLU activation, before being passed to a hidden layer of dimensionality 512
with no activation to provide the final representation $h_p$.

\paragraph{Data Source 1: Reddit.}
For training text representations we use a Reddit dataset similar to \newcite{AlRfou:2016arxiv}. Our dataset is large and provides natural conversational structure: all Reddit data from January 2015 to December 2018, available as a public BigQuery dataset, span almost 3.7B comments \cite{Henderson:2019convai}. We preprocess the dataset to remove uninformative and long comments by retaining only sentences containing more than 8 and less than 128 word tokens. After pairing all comments/contexts $c$ with their replies $r$, we obtain more than 727M context-reply $(c,r)$ pairs for training, see Figure~\ref{fig:arch}.

\paragraph{Data Source 2: Yelp.}
Once the text encoding sub-networks are trained, a photo encoder is learned on top of a pretrained MobileNet CNN, using data taken from the Yelp Open dataset:\footnote{\url{https://www.yelp.com/dataset}} it contains around 200K photos and their captions. Training of the multi-modal sub-network then maximises the similarity of captions encoded with the response encoder $h_r$ to the photo representation $h_p$. As a result, we can compute the score of a photo given a context using the cosine similarity of the respective vectors. A photo will be scored highly if it looks like its
caption would be a good response to the current context.\footnote{Note that not all of the Yelp dataset has captions, which is why we need to learn the photo representation. If a photo caption is available, then the response vector representation of the caption is averaged with the photo vector representation to compute the score. If a caption is not available at inference time, we use only the photo vector representation.}

\paragraph{Index of Responses.}
The Yelp dataset is used at inference time to provide text and photo candidates to display to the user at each step in the conversation. Our restaurant search is currently deployed separately for each city, and we limit the responses to a given city. For instance, for our English system for Edinburgh we work with 396 restaurants, 4,225 photos (these include additional photos obtained using the Google Places API without captions), 6,725 responses created from the structured information about restaurants that Yelp provides,
converted using simple templates to sentences of the form such as ``Restaurant X accepts credit cards.'', 125,830 sentences extracted from online reviews.

\paragraph{PolyResponse in a Nutshell.}
The system jointly trains two encoding functions (with shared word embeddings) $f(context)$ and
$g(reply)$ which produce encodings $h_c$ and $h_r$, so that the similarity $S(c,r)$ is high for all $(c,r)$ pairs from the Reddit training data and low for random pairs. The encoding function $g()$ is then frozen, and an encoding function $t(photo)$ is learnt which makes the similarity between a photo and its associated caption high for all \textit{(photo, caption)} pairs from the Yelp dataset, and low for random pairs. $t$ is a CNN pretrained on ImageNet, with a shallow one-layer DNN on top. Given a new context/query, we then provide its encoding $h_c$ by applying $f()$, and find plausible text replies and
photo responses according to functions $g()$ and $t()$, respectively. These should be responses that look like answers to the query, and photos that look like they would have captions that would be answers to the provided query. 

At inference, finding relevant candidates given a context reduces to computing $h_c$ for the
context $c$ , and finding nearby $h_r$ and $h_p$ vectors. The response vectors can all be pre-computed, and the nearest neighbour search can be further optimised using standard libraries such as Faiss \cite{Faiss:2017arxiv} or approximate nearest neighbour retrieval \cite{Malkov:2016arxiv}, giving an efficient search that scales to billions of candidate responses.

The system offers both voice and text input and output. Speech-to-text and text-to-speech conversion in the PolyResponse system is currently supported by the off-the-shelf Google Cloud tools.\footnote{https://cloud.google.com/speech-to-text/; https://cloud.google.com/text-to-speech/}

\section{Dialogue Flow}
\label{s:flow}
The ranking model lends itself to the one-shot task of finding the most relevant responses in a given context. However, a restaurant-browsing system needs to support a dialogue flow where the user finds a restaurant, and then asks questions about it. The \textit{dialogue state} for each search scenario is represented as the set of restaurants that are considered relevant. This starts off as \textit{all the restaurants} in the given city, and is assumed to monotonically decrease in size as the conversation progresses until the user converges to a single restaurant. A restaurant is only considered valid in the context of a new user input if it has relevant responses corresponding to it. This flow is summarised here:

\vspace{1.3mm}
\noindent S1. Initialise $R$ as the set of all restaurants in the city. Given the user's input, rank all the responses in the response pool pertaining to restaurants in $R$.

\vspace{1.3mm}
\noindent S2. Retrieve the top $N$ responses $r_1, r_2, \ldots, r_N$ with corresponding (sorted) cosine similarity scores: $s_1 \geq s_2 \geq \ldots \geq s_N$.

\vspace{1.3mm}
\noindent S3. Compute probability scores $p_i \propto \exp(a \cdot s_i)$ with $\sum_{i=1}^N p_i$, where $a>0$ is a tunable constant.

\vspace{1.3mm}
\noindent S4. Compute a score $q_e$ for each restaurant/entity $e \in R$, $q_e = \sum_{i: r_i \in e} p_i$.

\vspace{1.3mm}
\noindent S5. Update $R$ to the smallest set of restaurants with highest $q$ whose $q$-values sum up to more than a predefined threshold $t$.

\vspace{1.3mm}
\noindent S6. Display the most relevant responses associated with the updated $R$, and return to S2.

\vspace{1.3mm}
If there are multiple relevant restaurants, one response is shown from each. When only one restaurant is relevant, the top $N$ responses are all shown, and relevant photos are also displayed. The system does not require dedicated understanding, decision-making, and generation modules, and this dialogue flow does not rely on explicit task-tailored semantics. The set of relevant restaurants is kept internally while the system narrows it down across multiple dialogue turns. A simple set of predefined rules is used to provide a templatic spoken system response: e.g., an example rule is ``One review of $e$ said
$r$'', where $e$ refers to the restaurant, and $r$ to a relevant response associated with $e$. Note that while the demo is currently focused on the restaurant search task, the described ``narrowing down'' dialogue flow is generic and applicable to a variety of applications dealing with similar entity search.

The system can use a set of intent classifiers to allow resetting the dialogue state, or to activate the separate restaurant booking dialogue flow. These classifiers are briefly discussed in \S\ref{s:other}. 

\begin{figure*}[t]
    \centering
    \begin{subfigure}[t]{0.48\linewidth}
        \centering
        \includegraphics[width=1.0\linewidth]{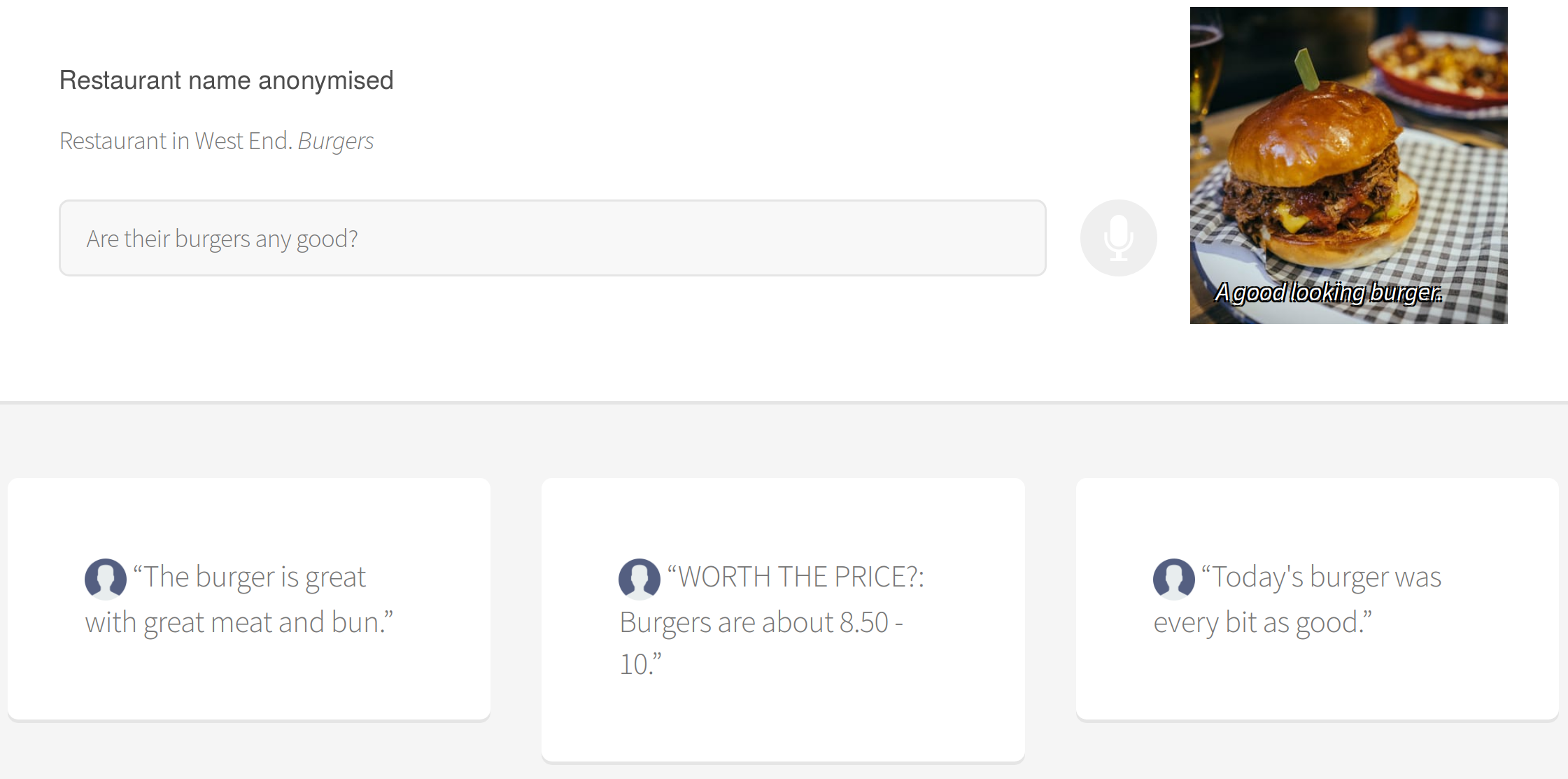}
        \caption{English system. City: Edinburgh.}
        \label{fig:elmo}
    \end{subfigure}%
    \hspace{1em}
    \begin{subfigure}[t]{0.48\textwidth}
        \centering
        \includegraphics[width=1.00\linewidth]{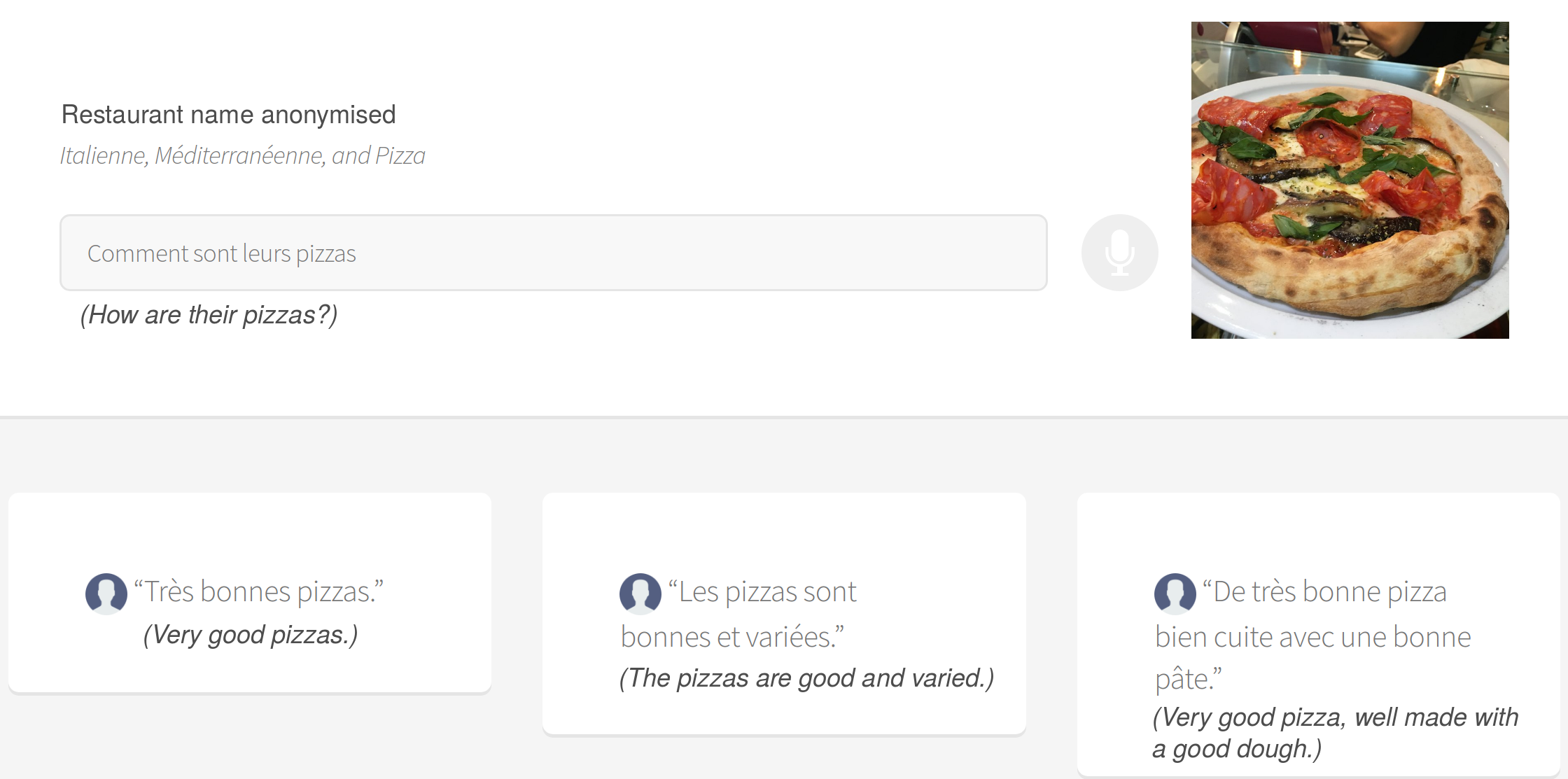}
        \caption{French system. City: Paris.}
        \label{fig:use}
    \end{subfigure}
    \begin{subfigure}[t]{0.48\textwidth}
        \centering
        \includegraphics[width=1.00\linewidth]{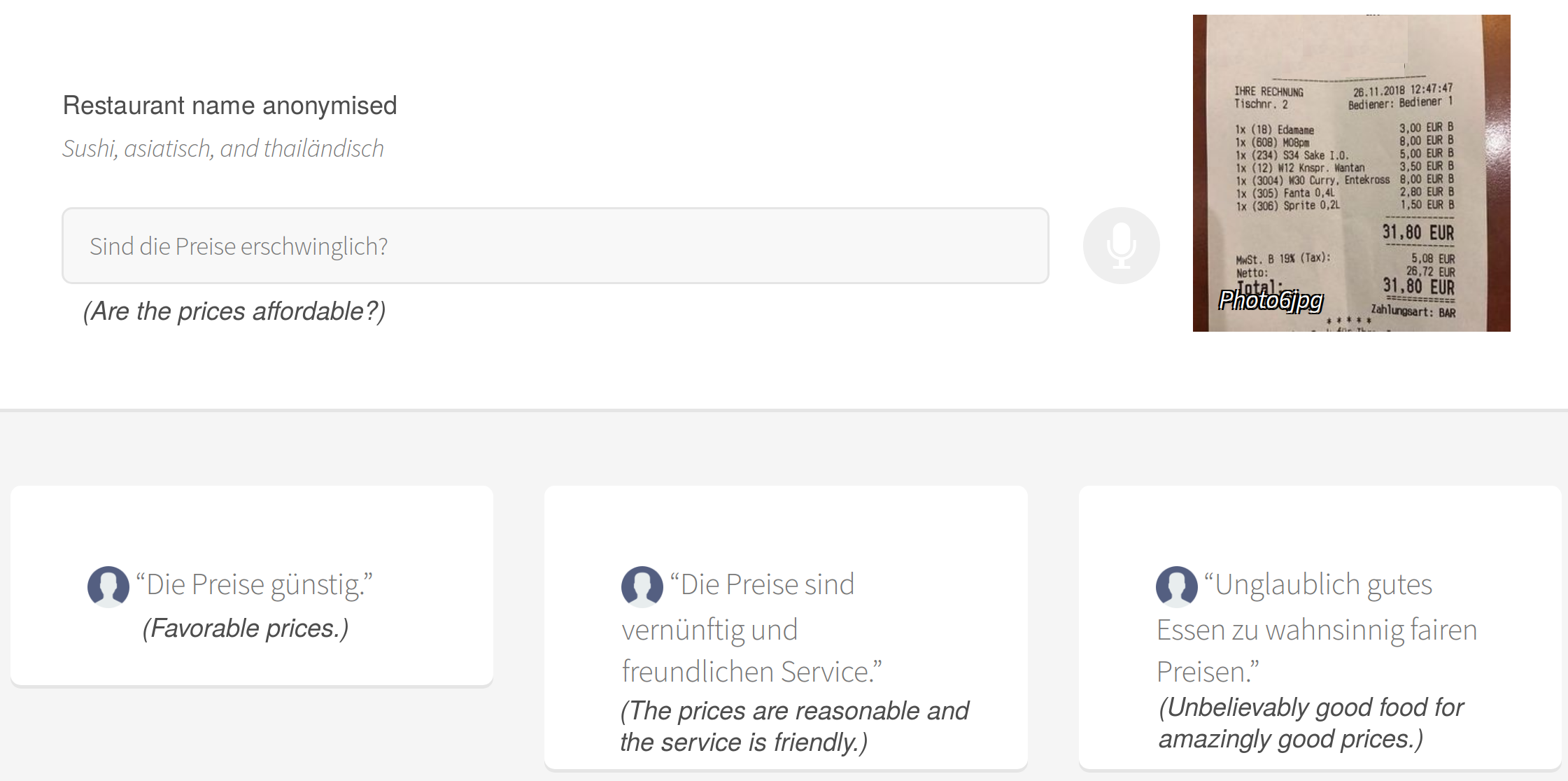}
        \caption{German system. City: Berlin.}
        \label{fig:nonfinetuned}
    \end{subfigure}%
    \hspace{1em}
        \begin{subfigure}[t]{0.48\textwidth}
        \centering
        \includegraphics[width=1.0\linewidth]{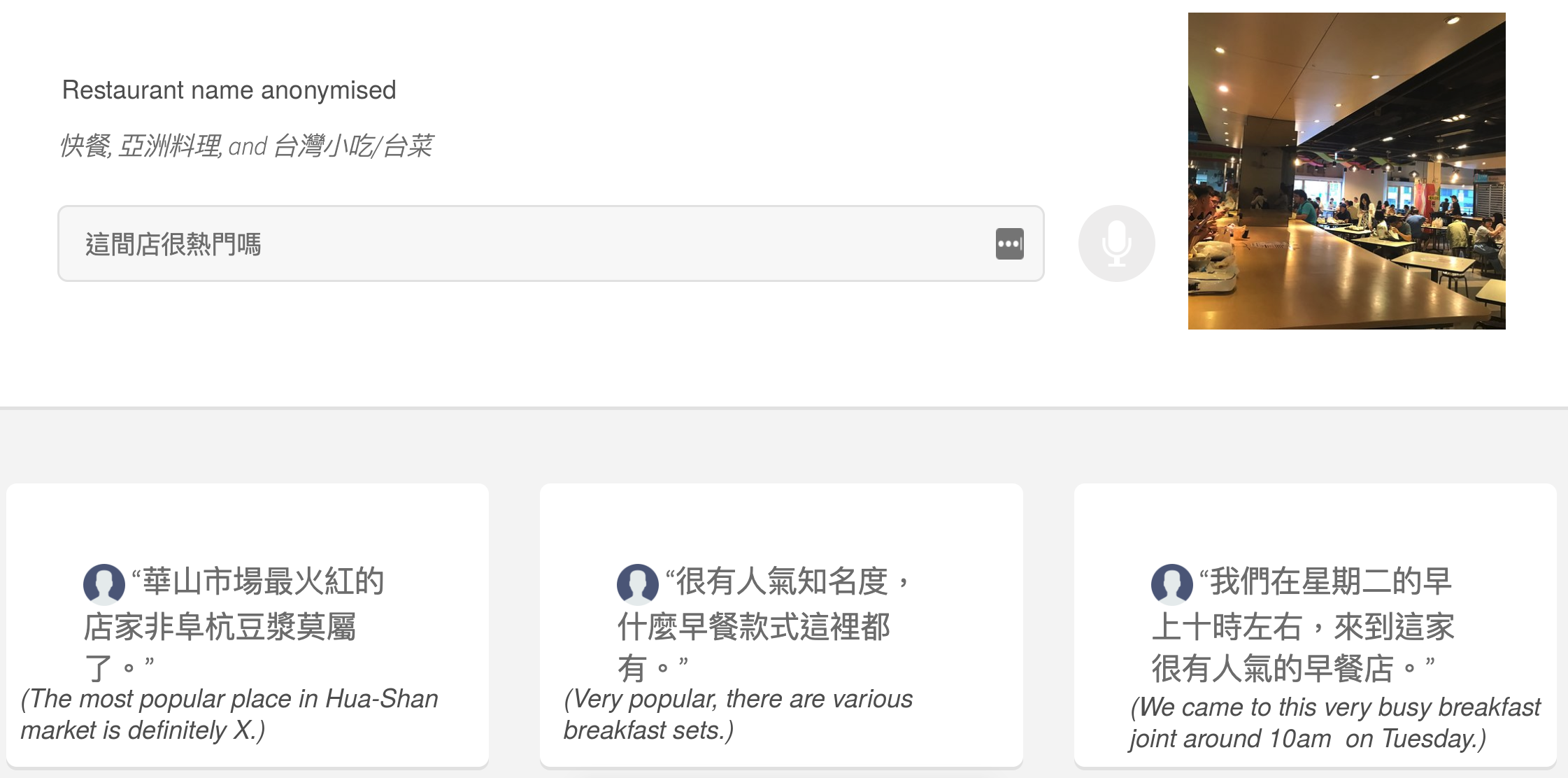}
        \caption{Mandarin system. City: Taipei.}
        \label{fig:finetuned}
    \end{subfigure}
    \vspace{-0.5mm}
	\caption{Snapshots of the PolyResponse demo system for restaurant search in four different languages. Restaurant names are anonymised. Translations of non-English sentences are provided in parentheses; they are not part of the system output. The output also comprises relevant photos associated with the current restaurant.}
\vspace{-2.5mm}
\label{fig:snapshots}
\end{figure*}

\section{Other Functionality}
\label{s:other}
\begin{figure}[!t]
\centering
    \includegraphics[width=0.99\linewidth]{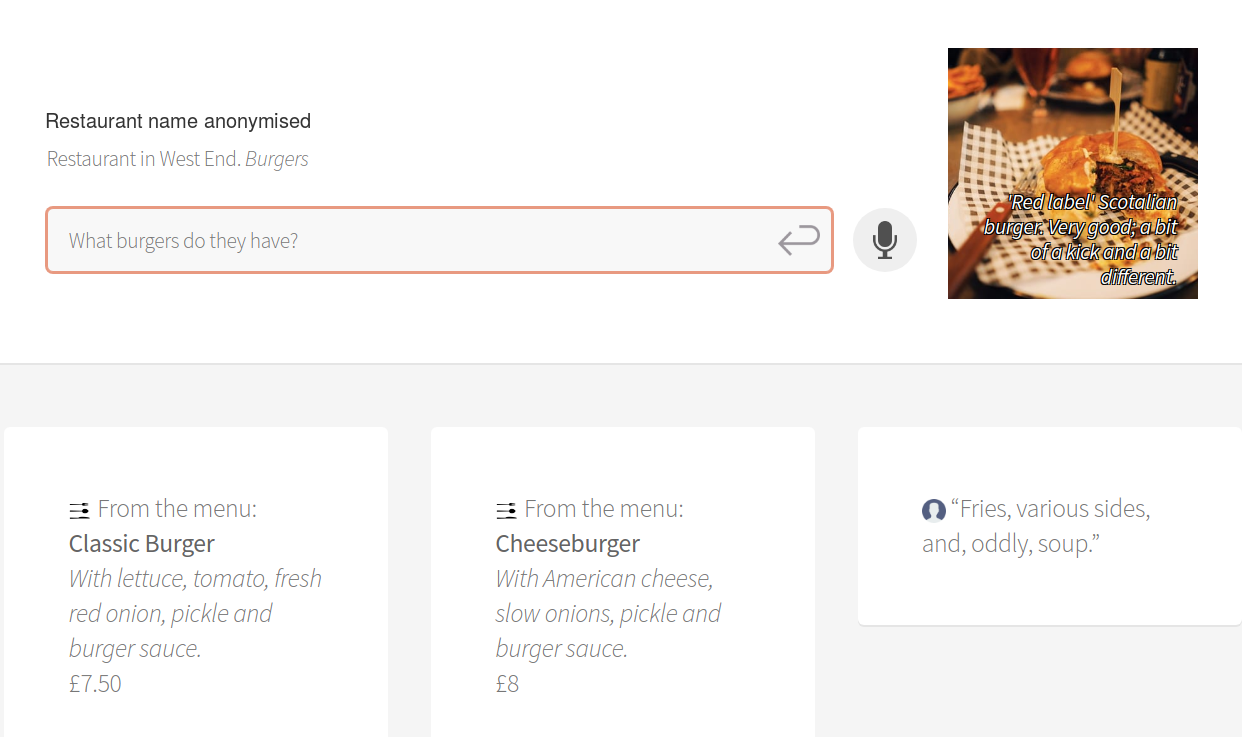}
	\caption{An example showing how the system can retrieve parts of the menu as responses to the
current user utterance (if they are relevant to the utterance).}
\vspace{-10.5mm}
\label{fig:menu}
\end{figure}

\paragraph{Multilinguality.}
The PolyResponse restaurant search is currently available in 8 languages and for 8 cities around the world: English (Edinburgh), German (Berlin), Spanish (Madrid), Mandarin (Taipei), Polish (Warsaw), Russian (Moscow), Korean (Seoul), and Serbian (Belgrade). Selected snapshots are shown in Figure~\ref{fig:snapshots}, while we also provide videos demonstrating the use and behaviour of the systems at: \url{https://tinyurl.com/y3evkcfz}. A simple MT-based translate-to-source approach at inference time is currently used to enable the deployment of the system in other languages: 1) the pool of responses in each language is translated to English by Google Translate beforehand, and pre-computed encodings of their English translations are used as representations of each foreign language response; 2) a provided user utterance (i.e., context) is translated to English on-the-fly and its encoding $h_c$ is then learned. We plan to experiment with more sophisticated multilingual models in future work.

\paragraph{Voice-Controlled Menu Search.}
An additional functionality enables the user to get parts of the restaurant menu relevant
to the current user utterance as responses. This is achieved by performing an additional ranking step of available menu items and retrieving the ones that are semantically relevant to the user
utterance using exactly the same methodology as with ranking other responses. An example of this functionality is shown in Figure~\ref{fig:menu}.

\paragraph{Resetting and Switching to Booking.}
The restaurant search system needs to support the discrete actions of restarting the conversation (i.e., resetting the set $R$), and should enable transferring to the slot-based table booking flow. This is achieved using two binary intent
classifiers, that are run at each step in the dialogue. These classifiers make use of the already-computed $h_c$ vector that represents the user's latest text. A single-layer neural net is learned on top of the $512$-dimensional encoding, with a ReLU activation and 100 hidden nodes.\footnote{Using the Reddit encoding has shown better generalisation when compared to models learned from scratch. This follows a recent trend where small robust classifiers are learned on pretrained large
models \cite{Devlin:2018arxiv}.} To train the classifiers, sets of 20 relevant paraphrases (e.g., ``Start again'') are provided as positive examples. Finally, when the system successfully switches to the booking scenario, it proceeds to the slot filling task: it aims to extract all the relevant booking information from the user (e.g., date, time, number of people to dine). The entire flow of the system illustrating both the search phase and the booking phase is provided as the supplemental video material.

\section{Conclusion and Future Work}
This paper has presented a general approach to search-based dialogue that does not rely on explicit semantic representations such as dialogue acts or slot-value ontologies, and allows for multi-modal responses. In future work, we will extend the current demo system to more tasks and languages, and work with more sophisticated encoders and ranking functions. Besides the initial dialogue flow from this work (\S\ref{s:flow}), we will also work with more complex flows dealing, e.g., with user intent shifts.

\bibliography{xx-refs}
\bibliographystyle{acl_natbib}

\label{sec:supplemental}

\end{document}